\documentclass[runningheads,a4paper]{llncs}

\usepackage[T1]{fontenc}
\usepackage{amsmath} 
\usepackage{amsfonts}
\usepackage{booktabs} 
\usepackage{placeins}
\usepackage{algorithm}
\usepackage{algorithmic}
\usepackage{graphicx}

\title{Tensor-based Fusion of Multi-modal Data for Predicting Cancer Risk Levels}

\author{Ali Anaissi$^{1,2}$, Yuanzhe Jia$^{1}$, Qishun Peng$^{1}$, Weidong Huang$^{2}$, Ali Braytee$^{2}$, Junaid Akram$^{1,2,3}$}

\institute{
University of Sydney, Australia \and
University of Technology Sydney, Australia \and
University of New South Wales, Australia \\
\email{ali.anaissi@uts.edu.au, yjia5612@uni.sydney.edu.au, qpen4596@uni.sydney.edu.au, weidong.huang@uts.edu.au, ali.braytee@uts.edu.au, junaid.akram@sydney.edu.au}
}

\begin{document}

\maketitle

\begin{abstract}
Cancer diagnosis and treatment increasingly rely on multi-modal diagnostic technologies that generate high-dimensional datasets. Although these datasets are rich in pathological information, they pose considerable challenges in interpretation, dimensionality reduction, and clinical applicability. Conventional machine learning approaches often fail to extract meaningful cross-modal features, thereby limiting predictive accuracy in patient risk stratification. To address these limitations, this paper proposes an unsupervised machine learning framework that integrates feature extraction, tensor decomposition, and clustering to improve cancer risk prediction. The framework uses tensor decomposition to fuse heterogeneous genomic and clinical data, enabling the extraction of latent, biologically meaningful features that outperform uni-modal feature sets. Multiple feature extraction strategies—Principal Component Analysis (PCA), truncated Singular Value Decomposition (tSVD), and Multidimensional Scaling (MDS)—are investigated, with hyperparameters optimized to preserve more than 95\% of variance across high-dimensional omics datasets. The extracted features are then clustered using K-means, K-median, Gaussian Mixture Model (GMM), DBSCAN, and hierarchical clustering. Comparative evaluations based on internal and external metrics identify the most effective clustering configurations. The framework is validated on LinkedOmics data comprising 11,158 patients across 32 cancer types. Five modalities are integrated: clinical annotation, DNA methylation, miRNA expression, RNA-seq profiles, and copy number variation. For patient risk stratification, clinically relevant attributes such as tumor purity, pathological stage, and survival outcomes serve as ground-truth indicators. Experimental results demonstrate the superiority of tSVD+GMM, which consistently outperforms other configurations across clinical benchmarks.
\end{abstract}


\section{Introduction}
The oncological landscape is profoundly shaped by the genetic basis of cancer, a heterogeneous group of diseases driven by disruptions across multiple biological levels~\cite{diaz2019tensor}. Such inherent genetic heterogeneity complicates diagnosis and treatment, contributing to diagnostic error rates of 15--45\% in U.S. clinical settings~\cite{esserman2013overdiagnosis}, which may lead to unnecessary surgical or chemotherapeutic interventions~\cite{bleyer2012effect}. Recent advances in gene sequencing and genomic data interpretation have enabled innovative methodologies that decipher the relationship between gene sequences and their functional outcomes. These advances include hybrid frameworks for RNA-seq imputation~\cite{anaissi2026hybrid}, as well as intelligent approaches for medical image recognition~\cite{feng2024integration}. Nevertheless, some cancers exhibit a chameleonic nature, presenting symptoms that mimic other diseases and often resulting in diagnostic inaccuracies~\cite{del2017missed}. In response, the medical community has pursued novel diagnostic paradigms, with AI technology offering promising avenues~\cite{liu2024ensemble}. However, a significant limitation of current AI models is their narrow focus on specific gene aspects, which introduces potential for substantial deviation and limits diagnostic accuracy~\cite{sjolander2008informative}.
To address these challenges, this paper introduces an unsupervised machine learning algorithm that integrates micro-level genomic and macro-level clinical patient data for the analysis of complex diseases, particularly cancer. The proposed approach identifies cancer patients through a comprehensive analysis of genomic and clinical data and further elucidates nuanced interactions among clinical outcomes to distinguish high- from low-risk patient categories. Its key contributions are fourfold: a. it establishes a paradigm for the holistic integration of micro- and macro-level patient data, setting a new precedent for comprehensive disease analysis; b. it advances methods for identifying cancer patients and stratifying them into distinct risk categories by examining clinical outcomes and effectively separating high- from low-risk groups; c. it achieves superior accuracy in cancer risk detection by exploiting data across multiple facets of cancer pathogenesis, thereby providing a robust predictive model; d. by amalgamating genomic and clinical insights, it offers a comprehensive predictive analysis model that can enhance prognostic assessment and support personalized therapeutic interventions.
The paper structure is organized as follows: Section~\ref{sec:relatedwork} reviews the contributions and limitations of existing research; Section~\ref{sec:prelim} presents the necessary preliminaries for the proposed model; Section~\ref{sec:method} details the methodology; Section~\ref{sec:experiment} reports the experimental results of the proposed model in terms of predictive capacity, stability, and applicability; and Section~\ref{sec:conclusion} summarizes the findings and outlines directions for future research.

\section{Related Work}
\label{sec:relatedwork}
Machine learning has become essential for predicting complex genome–disease relationships~\cite{liu2020imrm}. The introduction of Support Vector Machines (SVM) marked a major milestone in modeling genome–disease associations and demonstrated notable success for complex diseases~\cite{xu2011cancer}, as it computes similarities across heterogeneous data, including genomes and diseases~\cite{jiang2013vector}. However, SVM suffers from limitations such as the adverse effects of training with unlabeled samples and difficulty in obtaining accurate similarity measures under limited information. To mitigate these issues, semi-supervised methods have been developed, integrating disease-space predictions to improve genome–disease relationship models~\cite{chen2014semi}. Social network analysis techniques, including KATZ and CATAPULT, further enhance predictive capability by combining machine learning with network-based analysis: KATZ exploits known genome–disease associations, genome–genome functional similarity, and disease–disease similarity~\cite{zou2015prediction}, whereas CATAPULT uses clustering to construct an SVM classifier~\cite{wang2020data}. Tensor analysis has emerged as a new approach for identifying cancer-causing genes through tensor fusion and decomposition, including multi-domain data integration with joint SVD and CMF methods to improve functionality and reduce errors~\cite{pfeffer2019data}, as well as the CORCONDIA algorithm for determining the optimal number of PARAFAC components~\cite{bro2003new}. Models integrating multiple types of genetic data represent a new direction in disease prediction~\cite{zou2016similarity}, exemplified by the KBMF-MDI framework, which recalculates genome relationships and disease similarities to improve accuracy~\cite{lan2016predicting}, and MGDFMDE, which incorporates environmental factors through information fusion~\cite{peng2016framework}. Despite this progress, analyzing genomic or clinical data alone remains insufficient for diseases such as cancer, underscoring the need for comprehensive approaches that encompass all aspects of pathogenesis~\cite{robinson2012deep}. Because cancer development is influenced by both genetic and demographic factors, integrating diverse patient data is essential for cancer subtype models~\cite{bollig2016future}; tensor decomposition offers a promising solution for analyzing large datasets and enables high-throughput cancer subtype classification through sparse representation and efficient linear algebra operations~\cite{diaz2019tensor}.

\section{Preliminary}
\label{sec:prelim}
This section elucidates the preliminary methods underlying the proposed model, detailing the techniques employed for feature extraction, tensor decomposition and clustering to harness the predictive power of integrated data.

\subsection{Feature Extraction}

\subsubsection{Principal Component Analysis (PCA)}
is a fundamental dimensionality reduction technique that projects high-dimensional data into a lower-dimensional subspace while preserving as much variance as possible. Given a matrix $X \in \mathbb{R}^{n \times m}$, PCA first computes the covariance matrix $C = \frac{1}{m}XX^{\top}$, which characterizes the pairwise variances and correlations among features. The eigenvectors of $C$ define the principal component directions, and the corresponding eigenvalues quantify the variance captured by each direction. By sorting the eigenvalues in descending order and retaining the top $k$ eigenvectors, PCA constructs a projection matrix $V_k$. The original data are then transformed as $Z = XV_k$, yielding a $k$-dimensional representation that preserves the dominant structural information and maximizes the retained variance of the original data.

\subsubsection{Truncated Singular Value Decomposition (tSVD)}
factorizes a given matrix $X \in \mathbb{R}^{m \times n}$ as $X = U\Sigma V^T$, where $U \in \mathbb{R}^{m \times m}$ and $V \in \mathbb{R}^{n \times n}$ are orthogonal matrices, and $\Sigma \in \mathbb{R}^{m \times n}$ is a diagonal matrix whose diagonal entries are the singular values of $X$, typically ordered in descending magnitude. By retaining only the first $k$ singular values and their corresponding singular vectors, tSVD constructs the rank-$k$ approximation $X \approx U_k \Sigma_k V_k^T$, where $U_k$ and $V_k$ contain the leading $k$ left and right singular vectors, respectively. This operation projects the original data onto the $k$-dimensional subspace that captures the dominant variance and structural information, thereby reducing dimensionality while preserving the most important features of $X$.

\subsubsection{Multidimensional Scaling (MDS)}
is a distance-preserving dimensionality reduction technique that embeds data points into a lower-dimensional space by maintaining their pairwise dissimilarities, thereby enabling interpretable visualization and feature representation of high-dimensional data. Given a distance or dissimilarity matrix $\Delta = [\delta_{ij}]$, MDS seeks a low-dimensional embedding $X \in \mathbb{R}^{m \times n}$ such that the pairwise distances in $X$ approximate $\delta_{ij}$ as closely as possible, typically by minimizing a stress function.
MDS itself does not directly employ the Kronecker, Khatri-Rao, or Hadamard products; rather, these products provide the algebraic operations used downstream to construct and decompose the tensor formed from MDS-derived and other feature matrices. Their relationships can be summarized as follows. 

The Kronecker product is a general matrix product that generates all pairwise multiplicative interactions between two matrices. For $A \in \mathbb{R}^{I \times J}$ and $B \in \mathbb{R}^{K \times L}$, it is defined as

\[
A \otimes B =
\begin{bmatrix}
a_{11}B & a_{12}B & \cdots & a_{1J}B \\
a_{21}B & a_{22}B & \cdots & a_{2J}B \\
\vdots & \vdots & \ddots & \vdots \\
a_{I1}B & a_{I2}B & \cdots & a_{IJ}B
\end{bmatrix}
\in \mathbb{R}^{(IK) \times (JL)}.
\]

The Khatri-Rao product is the column-wise Kronecker product and is particularly useful in decomposition and alternating least squares. For $A \in \mathbb{R}^{I \times K}$ and $B \in \mathbb{R}^{J \times K}$, it is defined as

\[
A \odot B =
\begin{bmatrix}
a_1 \otimes b_1 & a_2 \otimes b_2 & \cdots & a_K \otimes b_K
\end{bmatrix}
\in \mathbb{R}^{(IJ) \times K}.
\]

The Hadamard product is the element-wise matrix product, operating on matrices of identical dimensions. For $A, B \in \mathbb{R}^{I \times J}$,

\[
A \ast B =
\begin{bmatrix}
a_{11}b_{11} & a_{12}b_{12} & \cdots & a_{1J}b_{1J} \\
a_{21}b_{21} & a_{22}b_{22} & \cdots & a_{2J}b_{2J} \\
\vdots & \vdots & \ddots & \vdots \\
a_{I1}b_{I1} & a_{I2}b_{I2} & \cdots & a_{IJ}b_{IJ}
\end{bmatrix}
\in \mathbb{R}^{I \times J}.
\]


\subsection{Tensor Decomposition}

\subsubsection{CANDECOMP/PARAFAC (CP) decomposition}
expresses a tensor into a sum of rank-one tensors, enabling the elucidation of multi-dimensional relationships within the data. This method converts a tensor into a matrix by reordering its elements. For a tensor $\chi \in \mathbb{R}^{I_1 \times I_2 \times \cdots \times I_N}$ with mode-$n$ matricization $X_{(n)}$, CP decomposition is defined as

\[
\chi \approx \sum_{r=1}^{R} a_r \circ b_r \circ c_r
\]

Thus, the tensor can be represented through three factor matrices, where each factor matrix is formed by combining the vectors of the rank-one components.

\[A  = \begin{bmatrix}
    a_{1} \quad a_{2} \quad \cdots \quad a_{R}\\
      \end{bmatrix}\]
\[B  = \begin{bmatrix}
     b_{1} \quad b_{2} \quad \cdots \quad b_{R}\\
      \end{bmatrix}\]
\[C  = \begin{bmatrix}
    c_{1} \quad c_{2} \quad \cdots \quad c_{R}\\
    \end{bmatrix}\]

Since an exact decomposition rarely exists for a specified number of components (referred to the rank), the goal of CP decomposition is to find the best possible approximation of the tensor. The decomposition is typically computed using the Alternating Least Squares (ALS) and the associated cost function can be expressed as

\begin{align*}
    \min\limits_{\widehat\chi}\| \chi- \widehat{\chi} \|with \widehat\chi = \sum_{r=1}^{R}\lambda_{r}a_{r}^{o}b_{r}^{o}c_{r} = [\![\lambda;A,B,C]\!]
\end{align*}

The principle of ALS is to iteratively update one factor matrix at a time while keeping the other matrices fixed, cycling through all factor matrices until the convergence criteria are met. By fixing two matrices, the optimization problem for the remaining matrix can be formulated as

\begin{align*}
    \min\limits_{\widehat{A}}\|X_{(1)}-\widehat{A}(C\odot B)^T\|_{F}
\end{align*}

Additionally, the key parameter in CP decomposition is the rank. To determine the optimal rank, CORCONDIA is used for hyperparameter tuning. CORCONDIA evaluates the core consistency, which can be computed as follows, where $F$ denotes the number of components.

\begin{align*}
    core consistency= 100\Biggl(1 -\frac{\sum_{d=1}^{F}\sum_{e=1}^{F}\sum_{f=1}^{F}(g_{def} - t_{def})^{2}}{F}\Biggl)
\end{align*}

\subsection{Clustering}

\subsubsection{K-means}
partitions a dataset into a predefined number $K$ of clusters by minimizing the within-cluster sum of squared distances. The algorithm begins by selecting $K$ initial centroids and assigning each data point to the cluster whose centroid is nearest under a given distance metric, typically Euclidean distance. Each centroid is then updated as the mean of the points assigned to its cluster, expressed as $a_i = \frac{1}{|c_i|}\sum_{x\in c_i} x$, where $a_i$ denotes the updated centroid, $c_i$ the current cluster, and $x$ a data point belonging to $c_i$. The assignment and update steps are repeated iteratively until the convergence criterion is satisfied.

\subsubsection{K-medians}
partitions a dataset into a predefined number $K$ of clusters by minimizing the sum of Manhattan distances between data points and their cluster centers. Like K-means, it alternates between an assignment step, in which each data point is assigned to the cluster whose center is nearest under the $L_1$ norm, and an update step, in which each cluster center is recomputed as the component-wise median of the points assigned to it. This median-based update, rather than the mean applied in K-means, makes K-medians much more robust to outliers and noise.

\subsubsection{Gaussian Mixture Model (GMM)}
represents the data as a mixture of $K$ multivariate Gaussian components, each characterized by a mean $\mu_k$, a covariance matrix $\Sigma_k$, and a mixing coefficient $\pi_k$. Clustering is performed via the Expectation-Maximization (EM) algorithm. In the expectation step, the posterior probability that a data point $x_i$ belongs to component $k$ is computed according to Bayes' rule, $\gamma_{ik} = \pi_k \mathcal{N}(x_i \mid \mu_k, \Sigma_k) / \sum_{j=1}^{K} \pi_j \mathcal{N}(x_i \mid \mu_j, \Sigma_j)$. In the maximization step, the model parameters are re-estimated by maximum likelihood using these posterior probabilities, with the mean updated as $\mu_k = \sum_{i} \gamma_{ik} x_i / \sum_{i} \gamma_{ik}$ and analogous updates applied to $\Sigma_k$ and $\pi_k$. The expectation and maximization steps are repeated iteratively until the convergence criterion is satisfied, after which each data point is assigned to the Gaussian component with the highest posterior probability.
Ultimately, GMM offers a probabilistic and flexible clustering framework that captures complex data distributions and enables soft assignments for downstream analysis.

\subsubsection{DBSCAN}
is a density-based clustering algorithm that defines clusters as dense regions of samples separated by regions of lower density. Given a radius parameter $\varepsilon$ and a minimum number of points $\mathit{minPts}$, a sample is regarded as a core point if its $\varepsilon$-neighborhood contains at least $\mathit{minPts}$ samples. Clusters are then formed by grouping core points and their density-reachable neighbors, whereas samples that are neither core nor border points are treated as noise. Consequently, DBSCAN can discover clusters of arbitrary shape without requiring the number of clusters to be specified in advance, while simultaneously identifying outliers in the dataset.

\subsubsection{Hierarchical Clustering}
constructs a nested clustering tree based on the similarity or dissimilarity between data points, where each leaf corresponds to an original data point and the root represents the entire dataset. The hierarchy can be built using two principal strategies: bottom-up agglomerative merging and top-down divisive splitting. Agglomerative clustering initially treats each data point as a singleton cluster and iteratively merges the closest pair of clusters according to a chosen linkage criterion until a desired number of clusters is obtained, whereas divisive clustering starts with all data points in a single cluster and recursively partitions larger clusters into smaller ones until the stopping criterion is met.

\subsubsection{Clustering in Quest (CLIQUE)}
is a subspace clustering algorithm designed for high-dimensional datasets, combining grid-based and density-based strategies to identify clusters of arbitrary shape while scaling efficiently to large data volumes. The algorithm first partitions each dimension of the data space into a predefined number of equal-width intervals, thereby discretizing the space into grid cells or units. A density threshold is then applied so that a unit is considered dense if the number of data points it contains exceeds this threshold. By identifying dense units across multiple subspaces and merging adjacent dense units into maximal connected regions, CLIQUE constructs a comprehensive embedding of the underlying cluster structure without requiring prior knowledge of the number of clusters.

\section{Methodology} 
\label{sec:method}

\begin{figure*}[ht]
    \centering
    \includegraphics[width=\textwidth]{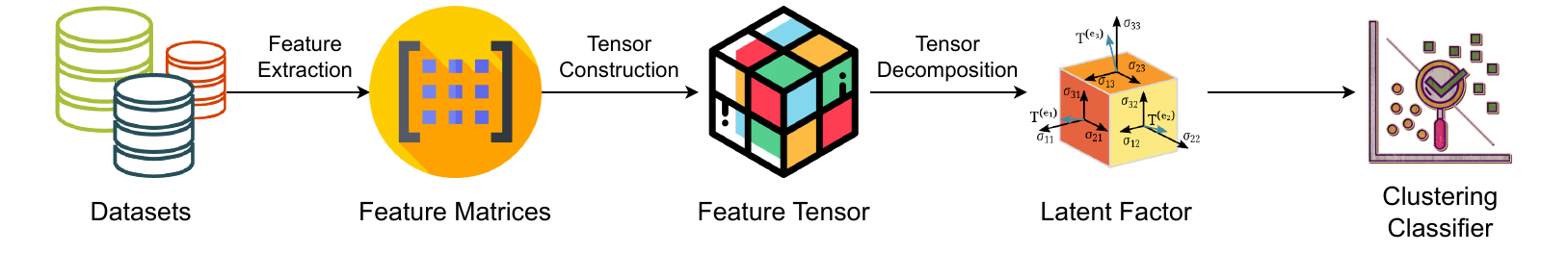}
    \caption{The architecture of the proposed model.}
    \label{fig:model}
\end{figure*}

Our proposed model follows a systematic pipeline comprising feature extraction, tensor decomposition, and clustering-based classifier construction, as illustrated in Figure~\ref{fig:model}, where feature matrices are derived from the datasets described in Section~\ref{sec:dataset}. Data preprocessing, including the removal of attributes with missing values, is first performed to enhance the accuracy and reliability of cancer prediction. Because the resulting feature matrices vary substantially in dimensionality---823, 20,155, 24,776, and 335,854---and the largest matrix still exceeds 310,000 dimensions after preprocessing, direct tensor decomposition and clustering are computationally impractical; dimensionality reduction is therefore essential for high-dimensional data. To this end, PCA, tSVD, and MDS are employed to project all matrices into a common low-dimensional space while preserving at least 95\% of the original variance, with 500 components identified as optimal across the genomic datasets. After reduction, each feature matrix has shape $m \times n$, where $m = 616$ denotes the number of patients and $n$ denotes the retained principal components. These reduced matrices are then stacked into a third-order tensor whose dimensions correspond to patients, features, and methods or modalities, with the patient dimension serving as the reference. Specifically, for two feature representations $x$ and $y$ of the same patient $p$, a dual-feature interaction matrix can be computed as $P = \mathrm{interact}(x[i], y[j])$. The tensor is constructed using a stacked model approach, in which different matrices are directly stacked, and the stacking order has negligible impact on the results. CP decomposition is subsequently applied, with the rank selected via CORCONDIA, which identifies rank 2 as optimal; this yields three factor dimensions---patients, features, and methods---and the patient dimension is selected for downstream analysis. In this dimension, multiple clustering classifiers are evaluated, including partition-based, density-based, model-based, grid-based, and hierarchical algorithms. After an initial filtering step, K-means, K-medians, DBSCAN, GMM, CLIQUE, and hierarchical clustering are considered, with K-means, K-medians, and GMM emerging as the primary partition- and model-based methods, and DBSCAN and CLIQUE representing density- and grid-based approaches, respectively. Clustering performance is assessed using both internal metrics, such as the Silhouette score and Calinski-Harabasz index, and external metrics, such as the Fowlkes-Mallows index and accuracy, against clinically relevant labels including tumor purity, pathologic T stage, pathologic N stage, and pathologic stage.

\section{Experiments}
\label{sec:experiment}

\subsection{Dataset}
\label{sec:dataset}
The data used in this study were obtained from LinkedOmics\footnote{\url{http://linkedomics.org}}, a comprehensive repository providing multi-omics and clinical data across a broad spectrum of cancer types. Specifically, we leverage datasets covering 32 cancer types and comprising 11,158 patients from the CGA project and 10 CPTAC cancer cohorts~\cite{vasaikar2018linkedomics}. Focusing on Breast Invasive Carcinoma, our analysis integrates five data subsets: clinical annotation for 1,097 patients with 19 attributes; CpG-site-level methylation data for 783 tumor samples with 335,854 attributes; normalized miRNA expression for 755 tumor samples with 823 attributes; normalized RNA-seq counts for 1,093 samples across 20,155 attributes; and gene-level copy number variations based on GISTIC2 log ratios for 1,080 samples with 24,776 attributes. This diverse and comprehensive collection, generated through varied techniques and measurements, provides the foundation for our investigation of Breast Invasive Carcinoma.

\subsection{Settings}

\begin{figure*}[ht]
    \centering
    \includegraphics[width=0.5\textwidth]{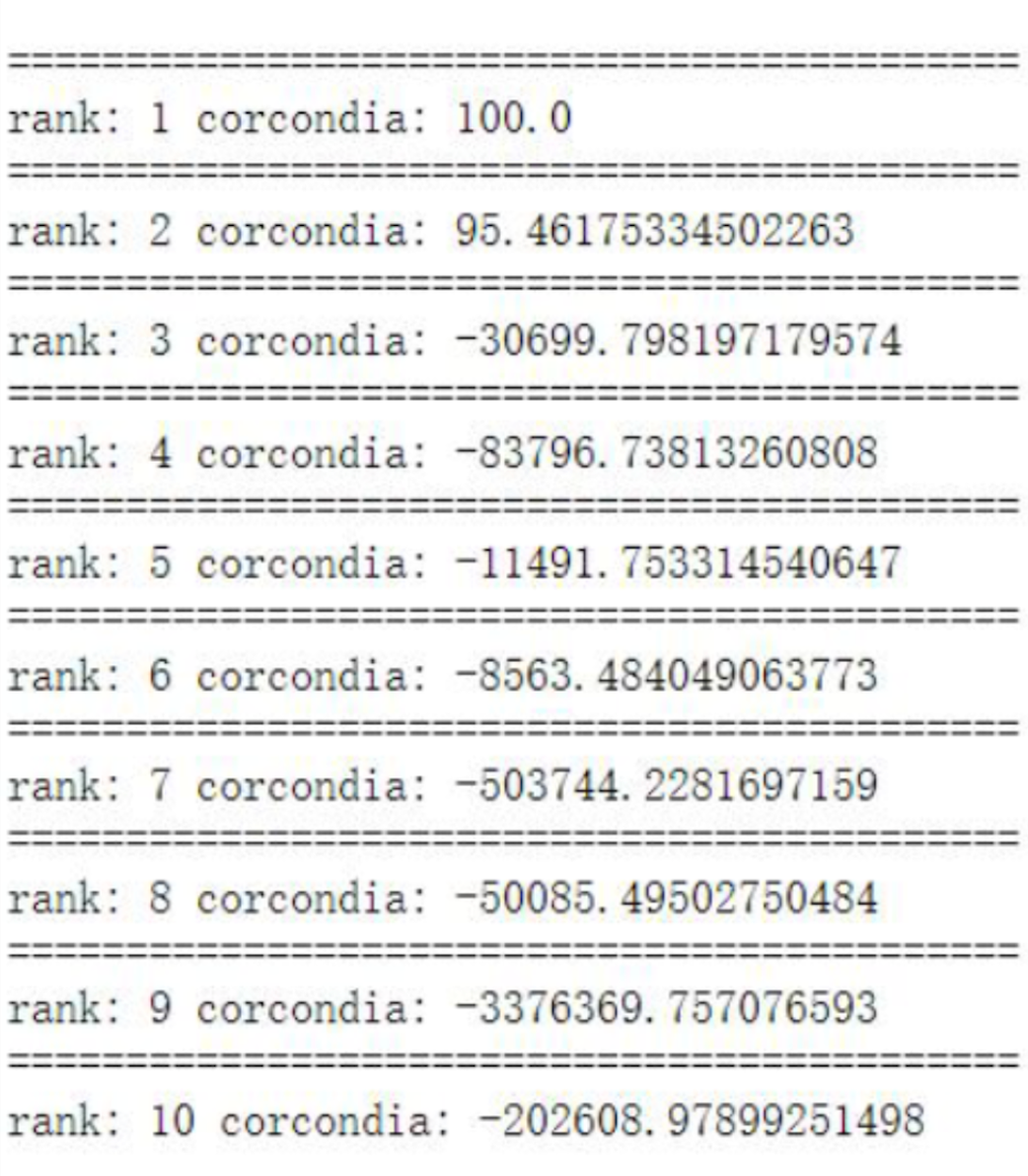}
    \caption{The selection the optimal rank via CORCONDIA.}
    \label{fig:corcondia}
\end{figure*}

We employ three principal feature extraction techniques: PCA, tSVD, and MDS. Each technique is characterized by a critical hyper-parameter--the number of components--which we optimized to capture a minimum of 95\% variance within the original data. Our analysis identified 500 as the optimal number of components, achieving the desired variance threshold across all genomic datasets.
The selection of the optimal rank for PARAFAC models, pivotal for tensor decomposition, was guided by the CORCONDIA metric. A systematic evaluation across ranks 1 to 10 identified rank 2 as optimal as demonstrated in Figure~\ref{fig:corcondia}, indicative of superior tensor decomposition performance.

\begin{table}[ht]
    \centering
    \caption{Comparative performance of hierarchical clustering configurations.}
    \label{tab:accuracy}
    \begin{tabular}{cccc}
    \hline
    Linkage & Distance Metric & Accuracy & Calinski-Harabasz \\
    \hline
    Complete & Euclidean & 0.015 & 168.010 \\
    Single & Euclidean & 0.003 & 7.022 \\
    Average & Euclidean & 0.027 & 207.776 \\
    Complete & Manhattan & 0.453 & 224.279 \\
    Single & Manhattan & 0.110 & 5.148 \\
    Average & Manhattan & 0.006 & 151.907 \\
    Complete & Cosine & 0.005 & 301.551 \\
    Single & Cosine & 0.453 & 0.047 \\
    Average & Cosine & 0.388 & 319.823 \\
    \hline
    \end{tabular}
\end{table}

Our exploration of clustering algorithms began with K-means, for which the optimal number of clusters was determined using evaluation metrics such as the Silhouette score and the Fowlkes-Mallows score; consistent results across multiple runs identified 2 as the ideal cluster number. For K-medians and GMM, the optimal cluster count mirrored that of K-means, also set to 2, with convergence typically achieved within 10 iterations. DBSCAN performance was calibrated by tuning its hyperparameters, namely minimum samples and $\epsilon$, under various distance metrics. Hierarchical clustering was parameterized by distance metrics and linkage criteria, further refining our clustering strategy. The comparative analysis, summarized in Table~\ref{tab:accuracy}, showed that the Manhattan distance with complete linkage yielded superior clustering outcomes.

\subsection{Evaluation and Discussion}

This section delineates the feature selection methodology, its rationale, and a comparative analysis of clustering algorithms. The evaluation phase aimed to assess the efficacy of the proposed model in stratifying patients into defined risk categories through deliberate segmentation of feature data aligned with predetermined cluster configurations.

\subsubsection{Feature Selection}
The clinical dataset was systematically examined to identify features instrumental in distinguishing high- from low-risk patient groups. Five attributes were selected based on their direct relevance to cancer prognosis and minimal missing values: tumor purity, pathologic T stage, pathologic N stage, pathologic stage, and overall survival. These attributes support binary risk classification, which is central to this study. Tumor purity was subjected to threshold analysis using the Jaccard coefficient~\cite{samanthula2016jaccard}, yielding an optimal cutoff of 0.75 that provides empirical support for the stratification strategy. Overall survival was excluded from the external evaluation criterion because discretizing it proved impractical for the present model. The relationship between tumor purity and patient risk classification offers visual justification for the adopted stratification approach.

\subsubsection{Evaluation Metrics}
The assessment framework combined internal and external metrics to evaluate clustering performance rigorously, using pathology T stage as the baseline. Internal metrics included the Silhouette score and Calinski-Harabasz score, which evaluate cluster cohesion and separation, while external metrics included the Fowlkes-Mallows index and accuracy, which measure concordance between clustering outcomes and clinical labels. Algorithms augmented with tSVD, specifically K-means, K-medians, and GMM, demonstrated superior Silhouette scores, indicating effective demarcation of patient groups. Among these, tSVD+K-means achieved the highest Calinski-Harabasz score, reflecting optimal inter-cluster dispersion. However, tSVD+GMM was most congruent with clinical categorizations and attained the highest accuracy, underscoring its superiority in clinical applicability and alignment with actual patient risk profiles.

\subsubsection{Comparative Analysis}
A key observation from the comparative analysis is that tSVD+GMM consistently outperformed its counterparts. For pathology N stage, all model combinations surpassed an accuracy benchmark of 0.5, with tSVD+GMM demonstrating superiority despite a marginally lower accuracy (0.661) than for pathology T stage. For pathologic stage, Silhouette scores ranged from 0.3 to 0.6, outperforming pathology N stage; notably, tSVD-enhanced algorithms, particularly tSVD+K-means, tSVD+K-medians, and tSVD+GMM, achieved Calinski-Harabasz scores above 1000. For tumor purity, its quantitative nature introduced a unique analytical dimension, and tSVD+GMM again proved effective, although its Fowlkes-Mallows score and accuracy slightly trailed those observed in the pathological stage assessments.

\subsubsection{Clustering Algorithms}
The exploration of DBSCAN and hierarchical clustering yielded mixed results. DBSCAN exhibited inconsistent cluster formation and struggled to meet the specified cluster count across different distance measures, underscoring its limitations for this dataset. Hierarchical clustering reached a maximum accuracy of 45\% but fell short of a robust performance benchmark, with numerous outcomes registering below 1\% accuracy. CLIQUE encountered challenges in label extraction from grid-based clustering outcomes, rendering it less efficacious for this dataset.

\subsubsection{Overview and Future Perspectives}
The integration of internal and external metrics provides comprehensive insight into the capabilities of the clustering models and underscores accuracy as the pivotal evaluation criterion. Across this multifaceted analysis, tSVD+GMM consistently demonstrated superior accuracy, robustness, and alignment with clinical categorizations in patient risk stratification. These findings position tSVD+GMM as a promising and clinically applicable approach in precision oncology, with the potential to advance cancer diagnostics and personalized treatment paradigms and to inform future translational and prognostic applications.

\section{Conclusion}
\label{sec:conclusion}
In this paper, we proposed an unsupervised machine learning framework for analyzing patient genomic and clinical data by integrating feature extraction, tensor decomposition, and clustering. The framework addresses the high-dimensional and heterogeneous nature of multi-modal biomedical data by applying PCA, tSVD, and MDS to project feature matrices into a common low-dimensional space, constructing a third-order tensor, and using CP decomposition to extract latent patient, feature, and method components. Clustering algorithms, including K-means, K-medians, GMM, DBSCAN, hierarchical clustering, and CLIQUE, were systematically evaluated using internal and external metrics. Experiments on LinkedOmics data showed that the tSVD+GMM pipeline consistently achieved superior accuracy, robustness, and alignment with clinically relevant risk indicators such as pathologic T and N stages and tumor purity, enabling effective stratification of high- and low-risk patients. These results demonstrate that tensor-based fusion of genomic and clinical data, combined with appropriate feature extraction and clustering, provides a reliable unsupervised approach for cancer risk prediction. 
Future work will extend the framework to additional cancer types and incorporate survival modeling and external validation to further advance precision oncology.

\bibliographystyle{splncs04}
\bibliography{ref}

\end{document}